\documentclass[11pt]{tea}

\usepackage[utf8]{inputenc}
\usepackage[T1]{fontenc}
\usepackage{hyperref}
\usepackage{xcolor}
\usepackage[most]{tcolorbox}

\definecolor{PromptGreen}{HTML}{1A531A}  
\definecolor{PromptGreenBg}{HTML}{F0F9F0}

\usepackage{subcaption}

\let\cite\citep

\title{Location-Aware Fine-Grained Representation Learning for Medical Vision Foundation Models}

\author[1,2]{Myeongkyun~Kang}
\author[1,2]{Yanting~Yang}
\author[1,2\dagger]{Xiaoxiao~Li}

\affiliation[1]{The University of British Columbia}
\affiliation[2]{Vector Institute}

\contribution{%
  $^\dagger$ Correspondence to Xiaoxiao Li at \href{mailto:xiaoxiao.li@ece.ubc.ca}{xiaoxiao.li@ece.ubc.ca}%
}

\abstract{
Fine-grained visual representations are essential for medical image analysis, particularly when diagnostically relevant evidence is subtle and spatially localized. Modern transformer-based medical vision encoders must therefore learn patch-level representations that are both clinically meaningful and spatially consistent. Without these properties, large vision-language models (LVLMs) operate on an ambiguous visual foundation, limiting their ability to generate clinically reliable and spatially grounded responses. However, existing training strategies for medical vision encoders rarely achieve both objectives. Image-text alignment provides clinically meaningful supervision primarily at the image level, leaving the spatial localization of diagnostic evidence weakly constrained. In contrast, self-supervised learning promotes spatial consistency but lacks the semantic supervision needed to distinguish visually similar yet clinically distinct regions. To address this gap, we present LoFi, a medical vision foundation model built on location-aware fine-grained representation learning. LoFi trains a vision encoder with a lightweight large language model under grounding and grounded captioning objectives. Because these objectives require predicting location from clinical text and vice versa, spatial consistency emerges without any explicit patch-level regularization. To enable training at scale, we construct MedG, a large-scale medical grounding dataset of 4.48M image-text-box triplets curated from 84 datasets spanning 7 modalities. Across phrase grounding, visual question answering, and region-based organ classification under perturbations, LoFi consistently outperforms general-purpose and medical vision foundation models as well as state-of-the-art LVLMs. Code is available at \url{https://github.com/myeongkyunkang/lofi-medg}.
}

\begin{document}
\maketitle

\section{Introduction}
Fine-grained representation is crucial in medical image analysis. In clinical practice, diagnosis proceeds in two steps: visual perception (inspecting the image) and cognition (interpreting the result)~\cite{krupinski2010current}. Perception is the more fundamental of the two, since downstream interpretation is upper-bounded by its accuracy~\cite{gray1978detection}. The same principle extends to medical large vision-language models (LVLMs): as diagnostic evidence is often subtle and spatially localized~\cite{huang2021gloria}, reliable reasoning hinges on a vision encoder that yields fine-grained representation, i.e., patch-level features that are both clinically meaningful and spatially consistent. Without these properties, the model is forced to reason on an ambiguous visual foundation, often producing spatially ungrounded responses~\cite{liu2026medical,asadi2026mirage}.

\begin{figure}[t]
    \centering
    \includegraphics[width=0.6\linewidth]{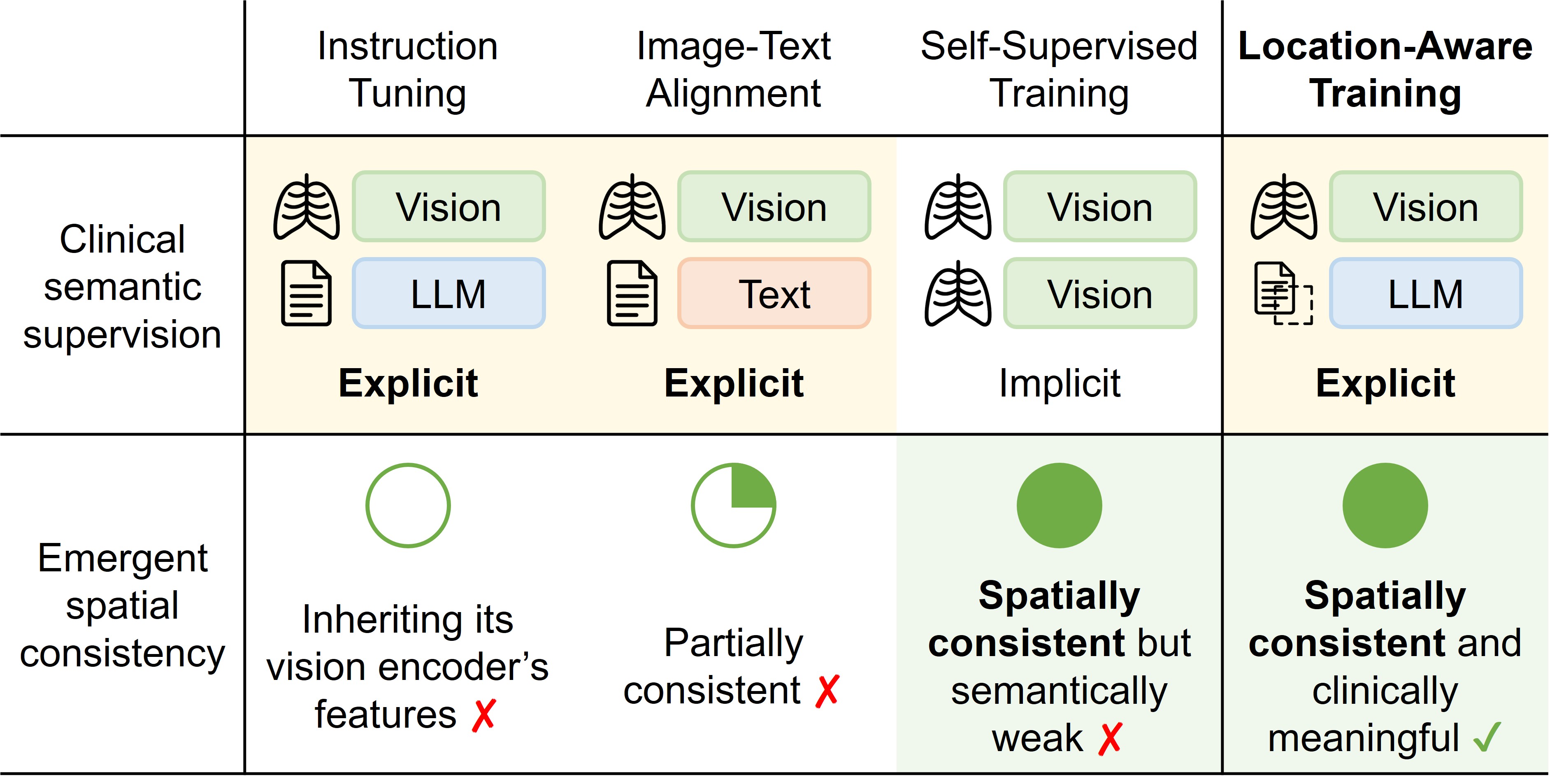}
    \caption{Three categories of existing medical vision foundation models and LoFi (ours). Encoders trained with explicit semantic supervision (e.g., the instruction-tuned Lingshu~\cite{xu2025lingshu} and the image-text-aligned MedSigLIP~\cite{sellergren2025medgemma}) lack spatial consistency, whereas those relying on self-supervision (e.g., RadioDINO~\cite{zedda2025radio}) are spatially consistent but semantically weak. Our location-aware training attains both, yielding patch-level features that are clinically meaningful and spatially consistent, and thus a stronger visual foundation for downstream medical LVLMs.}
    \label{fig_teaser}
\end{figure}

Current medical vision foundation models fall into three categories based on their training signal (Fig.~\ref{fig_teaser}). Instruction-tuned encoders~\cite{xu2025lingshu} are jointly trained within an LVLM, supervised explicitly through the language model. Image-text-aligned encoders~\cite{sellergren2025medgemma} are contrastively trained on paired image-text data, aligned explicitly but only at the image level. Self-supervised encoders~\cite{zedda2025radio} are trained without text (e.g., via self-distillation across augmented views), attaining spatial consistency but no semantic supervision. None of them, however, yields representations that are both clinically meaningful and spatially consistent. To illustrate this, we visualize each encoder's patch-level features at two complementary scopes. Within an image, PCA maps reveal spatial consistency: a spatially consistent encoder produces smooth patch-to-patch transitions. Within a region, PaCMAP~\cite{wang2021understanding} projections capture clinical meaningfulness: a clinically meaningful encoder separates patches from different organs.

\begin{figure}[t]
    \centering
    \begin{minipage}[t]{0.49\textwidth}
        \centering
        \includegraphics[width=\linewidth]{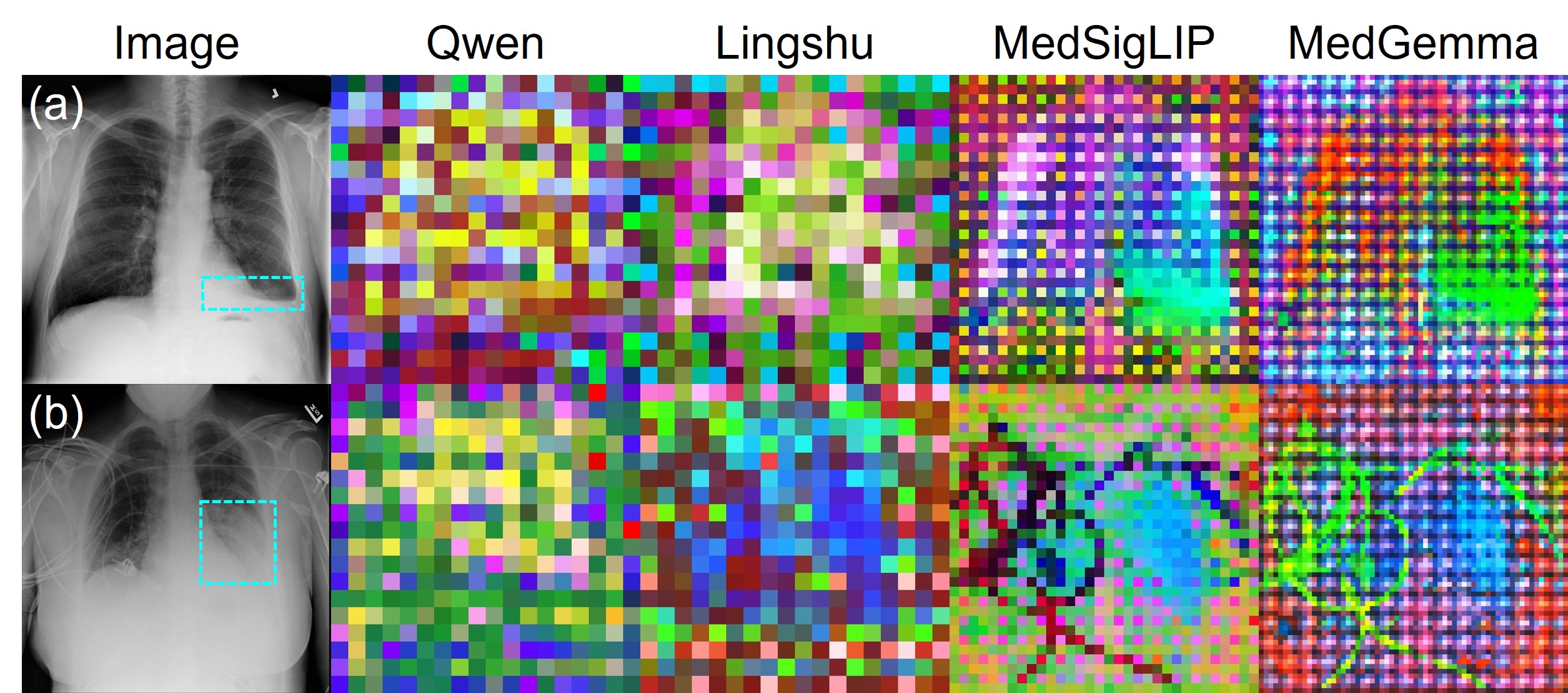}
        \captionof{figure}{PCA maps of patch-level features from Qwen~\cite{bai2025qwen25}, Lingshu, MedSigLIP, and MedGemma~\cite{sellergren2025medgemma}. Lingshu extends Qwen through instruction tuning on medical data, while MedGemma is built on the MedSigLIP vision encoder. For each image, the features are projected to three dimensions via PCA and mapped to RGB channels. The cyan box indicates the abnormal region.}
        \label{fig_pca}
    \end{minipage}
    \hfill
    \begin{minipage}[t]{0.49\textwidth}
        \centering
        \includegraphics[width=\linewidth]{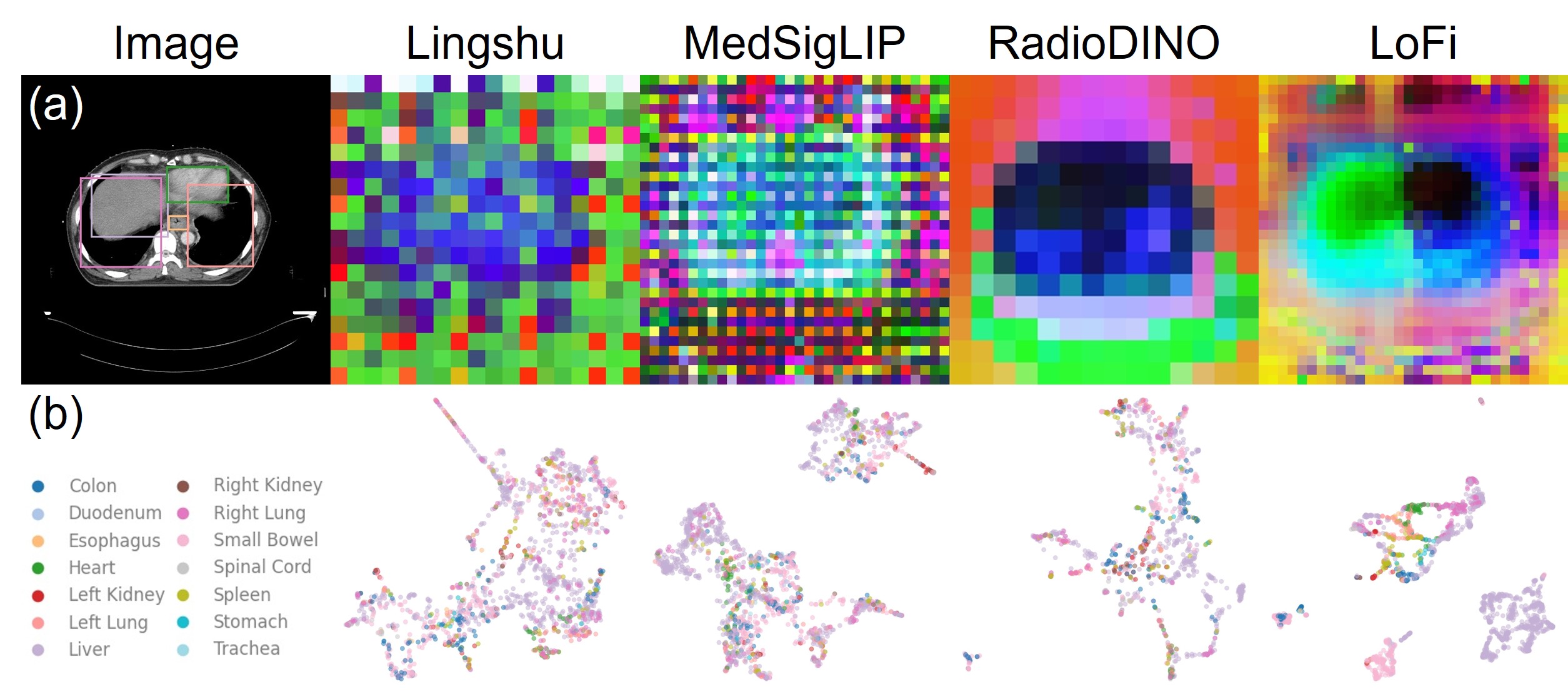}
        \captionof{figure}{PCA maps and PaCMAP~\cite{wang2021understanding} projections of patch-level features from Lingshu, MedSigLIP, RadioDINO, and LoFi (ours). For PaCMAP, patch-level features within the bounding boxes of 14 organ classes in SLAKE~\cite{liu2021slake} are projected to two dimensions, where each point is one patch-level feature.}
        \label{fig_projection}
    \end{minipage}
\end{figure}

As illustrated in Fig.~\ref{fig_pca}(a), the patch-level features of the instruction-tuned Lingshu~\cite{xu2025lingshu} are less discriminative in abnormal regions than those of the image-text-aligned MedSigLIP~\cite{sellergren2025medgemma}. This suggests that instruction tuning on large-scale medical data captures clinical semantics but does not necessarily enforce patch-level spatial consistency. Moreover, comparing Lingshu with its base LVLM (Qwen~\cite{bai2025qwen25}) and MedGemma~\cite{sellergren2025medgemma} with its vision encoder (MedSigLIP), we observe that instruction-tuned LVLMs largely inherit the patch-level properties of their vision encoders. Spatial consistency therefore needs to be addressed upstream, at the vision encoder rather than through instruction tuning.

While image-text-aligned encoders exhibit some emergent spatial consistency, such consistency remains insufficient. MedSigLIP encodes structures irrelevant to the downstream pulmonary task, such as tubes (Fig.~\ref{fig_pca}(b)). This is expected, as image-text alignment provides only image-level supervision, leaving spatial focus unconstrained. For the same reason, as shown in Fig.~\ref{fig_projection}(a), MedSigLIP's patch-level features are spatially inconsistent, yielding noisy PCA maps rather than smooth patch-to-patch transitions.

In contrast, the self-supervised RadioDINO~\cite{zedda2025radio} produces markedly more coherent PCA maps, highlighting the effectiveness of the self-distillation objective for spatial consistency. This consistency, however, arises purely from self-supervision and thus leaves semantics unconstrained; the encoder cannot distinguish organs of similar intensity, such as the liver and heart. The PaCMAP projections in Fig.~\ref{fig_projection}(b) confirm this: MedSigLIP and Lingshu mirror their noisy PCA maps, while RadioDINO fails to separate organs despite its high spatial consistency, leaving no existing encoder with both properties.

Existing medical vision encoders struggle to learn fine-grained representations for two reasons: (1) the scarcity of large-scale medical datasets with fine-grained annotations (e.g., bounding boxes), and (2) the lack of training strategies to exploit them. We address both. First, we construct \textbf{MedG}, a large-scale medical grounding dataset of 4.48M image-text-box triplets curated from 84 datasets across 7 modalities, surpassing prior work in both modality coverage (chest X-ray only~\cite{muller2025structured}) and scale~\cite{sheikh2026medrov}. Second, we propose a \textbf{Lo}cation-aware \textbf{Fi}ne-grained representation learning approach that yields \textbf{LoFi}, our medical vision foundation model. Prior approaches achieve spatial consistency through self-supervision or explicit patch-level regularization (e.g., a total variation loss), enforcing smoothness without regard to clinical semantics. Instead, we let spatial consistency emerge from clinical grounding: a grounding loss predicts bounding boxes from clinical text, while a grounded captioning loss reverses this, generating clinical text from a given region~\cite{wan2024locca}. Because localization yields spatially coherent patches and description makes them clinically meaningful, both properties emerge jointly without any explicit patch-level constraint (Fig.~\ref{fig_projection}). We optimize the two objectives autoregressively with a lightweight LLM~\cite{team2025gemma}, whose linguistic priors unify heterogeneous annotations (from single labels to report-derived phrases) and enable training at medical scale.
To evaluate the fine-grained representations captured by LoFi, we design a perception-centric phrase grounding task and a semantics-centric visual question answering (VQA) task, along with region-based organ classification under medical image perturbations.
Our main contributions are summarized as follows:
\begin{itemize}
    \item We construct MedG, a large-scale medical grounding dataset of 4.48M image-text-box triplets curated from 84 datasets across 7 modalities, with text annotations ranging from single labels to report-derived phrases.
    \item We introduce LoFi, a medical vision foundation model built on location-aware fine-grained representation learning. It trains a vision encoder with a lightweight LLM under grounding and grounded captioning objectives, allowing clinically meaningful and spatially consistent features to emerge without self-supervision or explicit patch-level regularization.
    \item Through extensive experiments on perception-centric, semantics-centric, and perturbation datasets, we show that LoFi consistently outperforms general-purpose and medical vision foundation models as well as state-of-the-art (SOTA) LVLMs.
\end{itemize}

A preliminary version of this work was provisionally accepted at MICCAI 2026~\cite{kang2026lofi}. That conference paper focused on chest X-ray analysis, training on MIMIC-CXR~\cite{johnson2019mimic} alone. This paper substantially extends the preliminary study in scope, analytical depth, and validation:
\begin{itemize}
    \item \textbf{New Large-Scale Multi-Modality Dataset:} The training data is expanded from a single chest X-ray source to MedG, covering 7 modalities across 84 curated datasets.
    \item \textbf{Added Qualitative Analysis of Fine-Grained Representations:} We qualitatively examine patch-level features across medical vision foundation models. This analysis reveals that neither image-text alignment nor self-supervision yields fine-grained representations that are both clinically meaningful and spatially consistent, motivating our location-aware objective.
    \item \textbf{Extended Broader Downstream Evaluation:} Evaluation is extended beyond chest X-ray grounding to perception-centric phrase grounding and semantics-centric VQA, benchmarking against general-purpose and medical vision foundation models.
    \item \textbf{Added Robustness Analysis:} We assess region-based organ classification under five clinically motivated perturbations, demonstrating greater robustness than self-supervised baselines.
    \item \textbf{Comparison with Additional SOTA Methods:} We add comparisons against recent medical LVLMs and an open-vocabulary detection model on both VQA and grounding.
\end{itemize}
\section{Related Works}
\subsection{Visual Representation Learning}
Vision encoders typically rely on either image-text alignment or purely self-supervised objectives. SigLIP~\cite{zhai2023sigmoid} introduces a sigmoid contrastive loss for image-text alignment, and SigLIP2~\cite{tschannen2025siglip} extends this with captioning, localization~\cite{wan2024locca}, and masked-prediction objectives to improve patch-level representations. LocCa~\cite{wan2024locca} adopts autoregressive grounding and grounded captioning as pretraining objectives, training from scratch on web-scale image-text pairs with pseudo box annotations. In contrast, DINO~\cite{caron2021emerging} relies on self-distillation between teacher and student networks across augmented views, and DINOv3~\cite{simeoni2025dinov3} extends this paradigm to large-scale training and further stabilizes patch-level representations through Gram anchoring. V-JEPA 2.1~\cite{mur2026v} predicts latent embeddings for both masked and visible patches over images and videos, leveraging intermediate-layer features for stronger patch-level representations.

To mitigate the domain gap, several methods specialize for medical imaging. MedSigLIP~\cite{sellergren2025medgemma} adapts SigLIP by continuing its sigmoid contrastive training on large-scale medical image-text pairs. For self-supervised learning, RadioDINO~\cite{zedda2025radio} adapts DINO to radiology via self-distillation on large-scale radiology data. On the segmentation side, MedSAM~\cite{ma2024segment} adapts Segment Anything~\cite{kirillov2023segment} by training on large-scale medical segmentation datasets, yielding an encoder that generalizes across diverse anatomical structures. These models, however, rely on image-level supervision, self-supervised objectives, or mask annotations, and thus capture either clinical semantics or spatial consistency, but rarely both. In contrast, LoFi is trained on large-scale image-text-box triplets, predicting location from clinical text and vice versa, thereby capturing both without any explicit patch-level constraint.

\subsection{Medical Large Vision-Language Models}
Several medical LVLMs have been developed by fine-tuning on carefully curated medical datasets. LLaVA-Med~\cite{li2023llava} and MedGemma~\cite{sellergren2025medgemma} integrate LLMs with domain-specific vision encoders, BiomedCLIP~\cite{zhang2025multimodal} and MedSigLIP respectively, and are fine-tuned on medical instruction-following data. More recent work instead adapts general-purpose LVLMs to the medical domain: Lingshu~\cite{xu2025lingshu} builds on Qwen2.5-VL~\cite{bai2025qwen25}, while Fleming~\cite{shu2025fleming} is based on InternVL3~\cite{zhu2025internvl3}. Although these models achieve strong performance on medical VQA benchmarks, recent studies show that they underutilize visual features and rely primarily on language priors during reasoning~\cite{liu2026medical,asadi2026mirage}. To improve generation reliability, the chest X-ray LVLMs CheXagent~\cite{chen2024chexagent}, MAIRA-2~\cite{bannur2024maira}, and RadVLM~\cite{deperrois2025radvlm} leverage bounding box annotations during instruction tuning to ground their outputs in localized visual evidence. Similarly, MedMO~\cite{deria2026medmo} applies reinforcement learning with a box-level reward to strengthen spatial grounding in general medical images. However, these post-training strategies remain constrained by the patch-level features inherited from the original vision encoder, indicating that the encoder itself must be strengthened upstream of the language model.
\begin{figure}[t]
    \centering
    \includegraphics[width=1.0\linewidth]{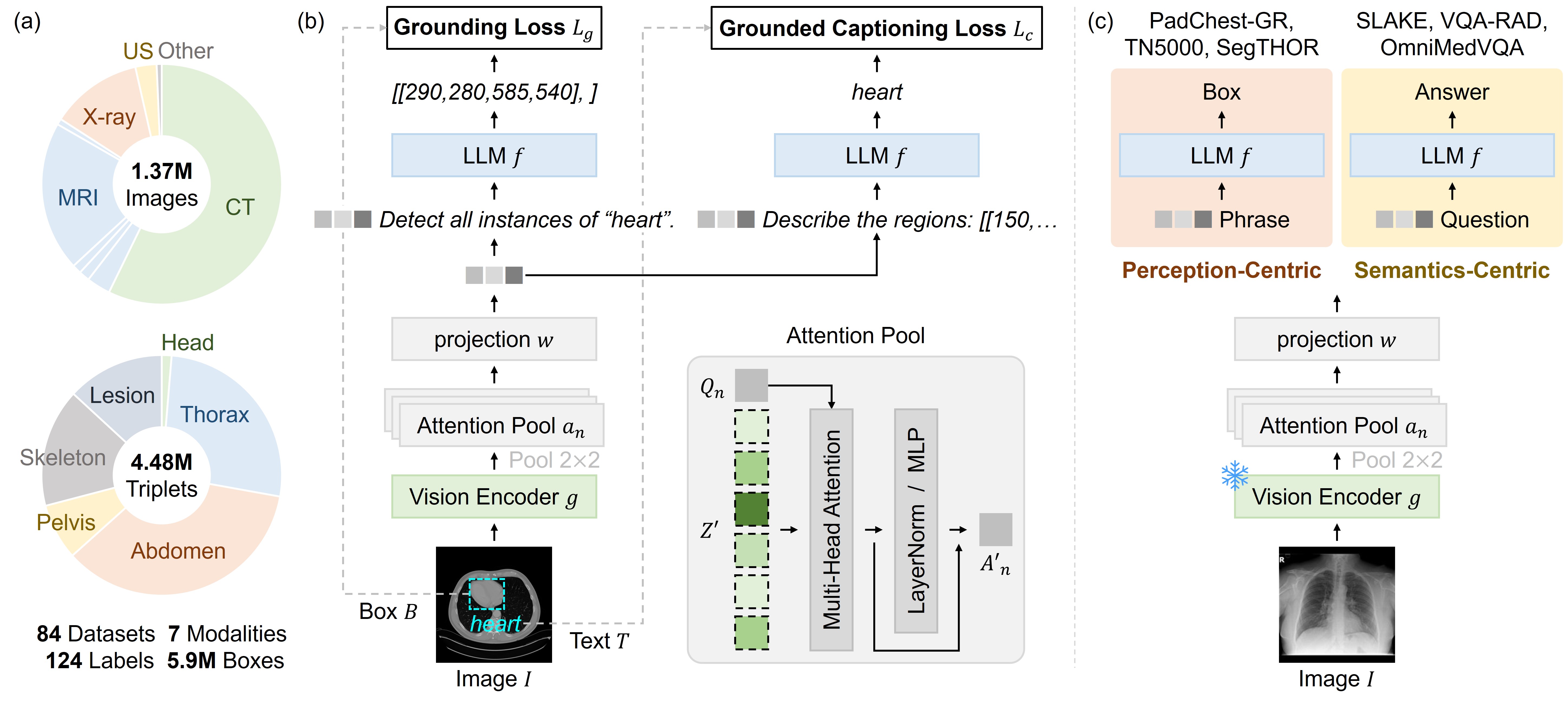}
    \caption{Overview of the location-aware fine-grained representation learning framework. (a) The MedG dataset. (b) LoFi training on image-text-box triplets $(I,T,B)$. (c) Fine-tuning of the vision encoder for perception-centric and semantics-centric tasks. The image $I$ is sequentially processed by the vision encoder $g$, attention pooling module $a_n$, projection layer $w$, and LLM $f$ to generate output tokens. The grounding loss $L_g$ and grounded captioning loss $L_c$ are computed using the bounding boxes $B$ and text $T$.}
    \label{fig_method}
\end{figure}

\section{Methodology}
\subsection{Medical Grounding Dataset}
We integrated publicly available preprocessed subsets from IMed-361M~\cite{cheng2025interactive}, TotalSegmentator CT~\cite{wasserthal2023totalsegmentator}, TotalSegmentator MRI~\cite{akinci2025totalsegmentator}, and MIMIC-CXR~\cite{johnson2019mimic}, spanning 84 medical datasets, to construct large-scale image-text-box triplets.
\textbf{IMed-361M} comprises 81 datasets, preprocessed into 2D images following the SA-Med2D-20M protocol~\cite{ye2023sa}. We normalized label names for consistency (e.g., mapping ``pulmonary nodule'' to ``lung nodule''), converted segmentation masks into text-box pairs, and excluded bounding boxes whose area was below 0.16\% of the image area.
\textbf{TotalSegmentator CT} contains 1,204 CT scans with 104 annotated anatomical structures, and \textbf{TotalSegmentator MRI} contains 616 MRI scans with 50 annotated anatomical regions. For CT, we applied Hounsfield unit (HU) windowing (window width 2000 HU, window level 0 HU) and extracted axial 2D slices from the 3D volumes, then converted the segmentation masks into text-box pairs as above. To improve label consistency across IMed-361M and TotalSegmentator CT/MRI, we removed laterality information (merging left and right labels) and collapsed indexed labels into their parent anatomical labels (e.g., mapping ``Vertebrae C1'' to ``Vertebrae'').
\textbf{MIMIC-CXR} contains 377,110 chest X-rays from 227,835 studies. We excluded non-frontal and low-quality images using~\cite{gaggion2024chexmask}, and obtained bounding box annotations from the positive A\textsuperscript{++}-rated text-box pairs across 153,602 studies in MIMIC-Ext-CXR-QBA~\cite{muller2025structured}. Duplicate boxes were merged via weighted box fusion.
All datasets used in this study were de-identified and released with ethical approval from the original providers. As summarized in Fig.~\ref{fig_method}(a), the imaging data span multiple modalities: CT, MRI, X-ray, ultrasound, fundus photography, dermoscopy, and endoscopy. MRI acquisitions included T1-weighted, T2-weighted, fluid-attenuated inversion recovery (FLAIR), gadolinium-enhanced T1-weighted, time-of-flight magnetic resonance angiography (TOF-MRA), as well as cardiac MRI. The harmonized label set comprises 124 categories, covering anatomical structures across the head, thorax, abdomen, pelvis, and skeleton, together with lesion labels.

\subsection{Location-Aware Fine-Grained Representation Learning}
\subsubsection{Architecture}
Given a dataset $D$ of image-text-box triplets $(I,T,B)$, we train a Vision Transformer (ViT)-based vision encoder $g$ by leveraging a lightweight LLM $f$, as illustrated in Fig.~\ref{fig_method}(b). For an input image $I$, the encoder $g$ produces patch-level features $Z = g(I)$ from its final Transformer layer. Following~\cite{team2025gemma}, we downsample $Z$ with $2 \times 2$ average pooling (stride 2) to reduce computational cost, i.e., $Z' = \operatorname{AvgPool}(Z)$. To construct visual tokens for the LLM, we employ $N$ attention pooling modules $\{a_n\}_{n=1}^{N}$. Each module $a_n$ applies multi-head attention (MHA) with a learnable probe $Q_n$ as the query and $Z'$ as the keys and values, yielding $A_n = \operatorname{MHA}(Q_n,Z',Z')$. The output is then passed through layer normalization (LayerNorm) and a multilayer perceptron (MLP) with a residual connection, resulting in $A'_n = A_n + \operatorname{MLP}\left(\operatorname{LayerNorm}(A_n)\right)$. Finally, we stack the pooled representations and project them into the LLM embedding space via a projection layer $w$, yielding $H = w\left(\operatorname{Stack}(A'_1,\ldots,A'_N)\right)$.

\subsubsection{Training}
To incorporate region-level supervision of fine-grained object locations, we train the model with a lightweight LLM using a grounding loss and a grounded captioning loss. Localizing a region from clinical text requires patch-level features to be spatially coherent within the region, while recovering the text from that region keeps those features clinically descriptive. Both losses use the paired text $T$ and bounding boxes $B$ to align textual content with the corresponding image regions. The text $T$ is either an anatomical label (e.g., ``heart'') from a set of 124 labels, or a phrase extracted from the corresponding MIMIC-CXR radiology report. Each bounding box in $B$ is represented by the coordinates of its top-left and bottom-right corners, $(x_{\min},y_{\min},x_{\max},y_{\max})$, normalized to integers in $[0,1000]$ (e.g., $[[290,280,585,540],]$). For training consistency, boxes paired with the same text $T$ are sorted in ascending order of $x_{\min}$. The sorted coordinates are then serialized into a string and tokenized in the same way as $T$.
With these region-level annotations, we optimize the parameters $\theta=\{g,a_{1:N},w,f\}$ under an autoregressive objective by minimizing the negative log-likelihood. Formally, the grounding loss is defined as
\begin{equation}
L_g
=
-\frac{1}{|B|}\sum_{i=1}^{|B|}
\log p_{\theta}
\left(
B_i \mid H, P_g, T, B_{<i}
\right),
\end{equation}
where $p_{\theta}(\cdot)$ denotes the conditional probability predicted by the autoregressive LLM, and $P_g$ is a fixed grounding prompt (e.g., \textit{``Detect all instances of''}). For grounded captioning, we adopt the same autoregressive formulation, except that the model predicts the text $T$ instead of the bounding boxes $B$. The grounded captioning loss is defined as
\begin{equation}
L_c
=
-\frac{1}{|T|}\sum_{i=1}^{|T|}
\log p_{\theta}
\left(
T_i \mid H, P_c, B, T_{<i}
\right),
\end{equation}
where $P_c$ is a fixed grounded captioning prompt (e.g., \textit{``Describe the regions:''}). 
The final training objective combines the two as $L = L_g + L_c$.

Our most closely related approach, LocCa~\cite{wan2024locca}, does not transfer to the medical domain: no detector generalizes across modalities well enough to pseudo-label at scale, and a decoder trained from scratch would not converge at MedG's scale. We therefore rely on curated ground-truth annotations rather than detector pseudo-labels, and replace the decoder with a pretrained lightweight LLM that supplies the linguistic prior that data scale alone cannot provide.

\subsection{Downstream Fine-Tuning}
As illustrated in Fig.~\ref{fig_method}(c), we evaluate the effectiveness of the trained vision encoder by fine-tuning it on downstream tasks such as phrase grounding and VQA. During this stage, the vision encoder is kept frozen, and only $a_{1:N}$, $w$, and $f$ are updated.
For phrase grounding, we employ a dataset $D_{pg}$ of image-text-box triplets $(I,T,B)$. The model is fine-tuned to predict the bounding boxes $B$ conditioned on the image $I$ and text phrase $T$, optimizing the grounding objective $L_g$.
For VQA, we use a dataset $D_{vqa}$ of image-question-answer triplets $(I,T_q,T_a)$. The model is fine-tuned to autoregressively generate the answer $T_a$ conditioned on the image $I$ and question $T_q$, without a fixed prompt template.
Beyond these main tasks, we extend $D_{pg}$ to region-based organ classification, which serves as our benchmark for evaluating robustness under perturbations. For this task, the model is fine-tuned to generate $T$ conditioned on the image $I$ and bounding boxes $B$, where $T$ begins with a special label token uniquely assigned to each organ, followed by the organ name (e.g., ``\texttt{<token1>}heart''). This token lets us analyze the entropy and shifts of the fine-tuned model's predicted probabilities under perturbation. We adopt the grounded captioning objective and minimize $L_c$ with the fixed prompt \textit{``Classify the regions:''}.

\subsection{Implementation}
We employed the pretrained SigLIP2-400M~\cite{tschannen2025siglip} with patch size 16 and input resolution 512 as the vision encoder $g$, and Gemma-3-270M~\cite{team2025gemma} as the lightweight LLM $f$. During training, we used $N=256$ attention pooling modules and set the maximum decoder length to 456 tokens, comprising 256 visual tokens and 200 generated text tokens. LoRA~\cite{hu2022lora} was applied to the query, key, value, and output projection layers, as well as the feed-forward layers, of $g$ with rank 16, and to the query and value layers of $f$ with rank 4. The model was trained for 10 epochs with a batch size of 64, where one epoch corresponds to 2M sampled instances. A cosine annealing learning rate schedule was used, with the learning rate decayed from $3\times10^{-4}$ to $3\times10^{-6}$. Weight decay was set to $1\times10^{-2}$, and the gradient norm was clipped to 1.0.
For downstream fine-tuning, we used $N=128$ attention pooling modules and set the maximum decoder length to 278 tokens, comprising 128 visual tokens and 150 generated text tokens. LoRA was applied to the query and value layers of $f$ with rank 4. The model was fine-tuned for 30 epochs with a batch size of 16. A cosine annealing learning rate schedule was used, with the learning rate decayed from $3\times10^{-4}$ to $3\times10^{-5}$. Weight decay was set to $1\times10^{-2}$, and the gradient norm was clipped to 1.0. We trained LoFi on a single NVIDIA B300 GPU and performed downstream fine-tuning on a single NVIDIA L40S GPU.
\begin{figure}[t]
    \centering
    \includegraphics[width=0.5\linewidth]{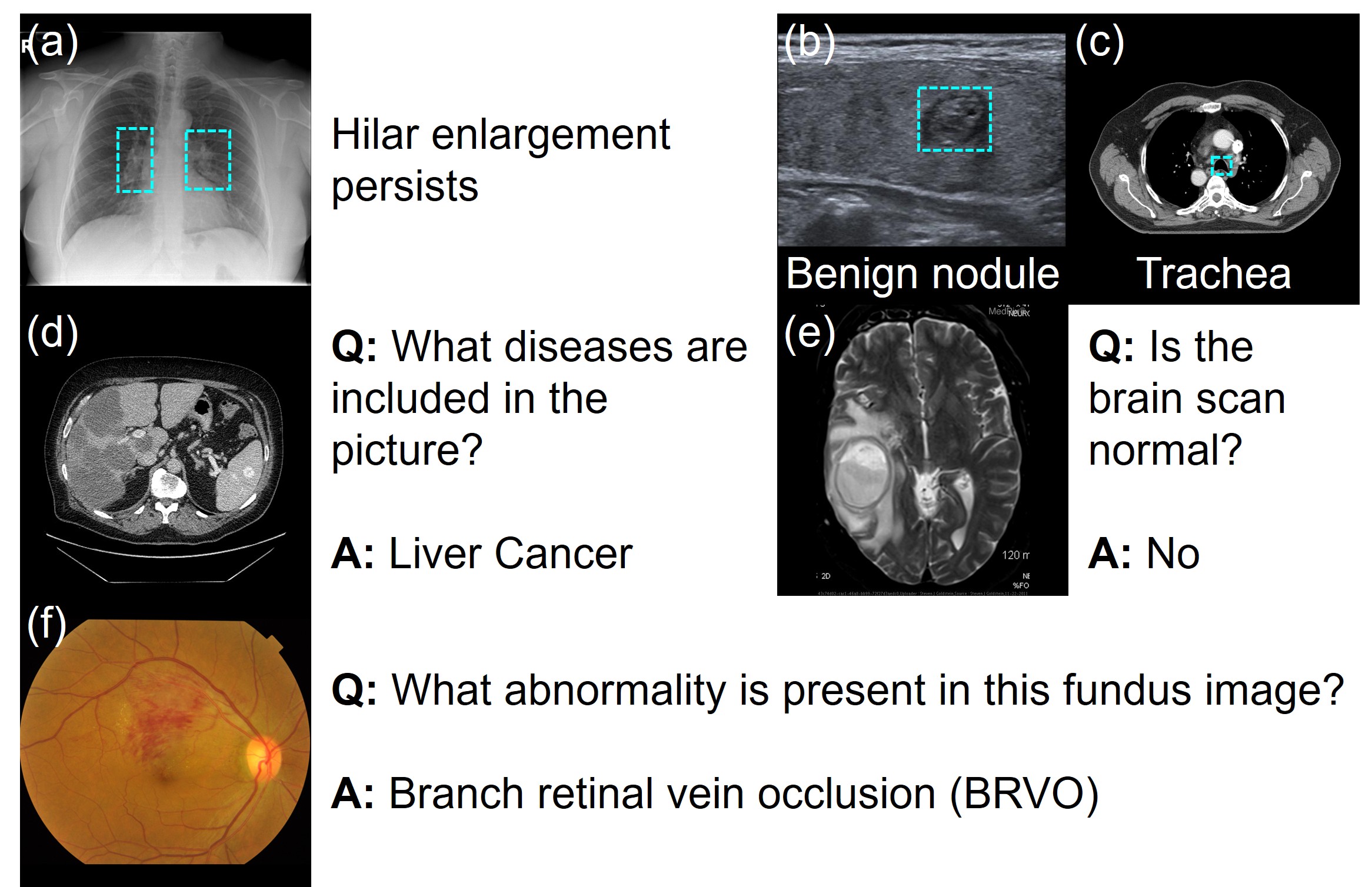}
    \caption{Examples of downstream datasets: (a) PadChest-GR, (b) TN5000, (c) SegTHOR, (d) SLAKE, (e) VQA-RAD, and (f) OmniMedVQA. The cyan box indicates the ground-truth bounding box corresponding to the text.}
    \label{fig_dataset}
\end{figure}

\section{Experiments}
We evaluate three aspects: \textit{perception}, using perception-centric datasets that require precise visual grounding; \textit{semantic understanding}, using semantics-centric datasets that require image-level interpretation; and \textit{robustness}, using a region-based organ classification dataset that contains both original and perturbed test images.
\subsection{Downstream Datasets}
\subsubsection{Perception-Centric Datasets}
Perception-centric datasets target tasks that require precise spatial grounding.
\textbf{PadChest-GR}~\cite{de2025padchest} consists of 4,555 chest X-rays paired with grounding reports that contain text-box annotations (see Fig.~\ref{fig_dataset}(a)). After excluding annotations with empty bounding boxes, we followed the official split, using 4,341 text-box pairs from 2,096 chest X-rays for training, 638 pairs from 308 X-rays for validation, and 1,238 pairs from 604 X-rays for testing.
\textbf{TN5000}~\cite{zhang2025tn5000} consists of 5,000 thyroid ultrasound images with biopsy-confirmed benign/malignant labels and bounding boxes (see Fig.~\ref{fig_dataset}(b)). We followed the official split, with 3,500 text-box pairs from 3,500 images for training, 500 from 500 images for validation, and 1,000 from 1,000 images for testing.
\textbf{SegTHOR}~\cite{lambert2020segthor} consists of 40 thoracic CT scans with ground-truth segmentation masks for four anatomical structures: esophagus, heart, trachea, and aorta (see Fig.~\ref{fig_dataset}(c)). We applied conventional soft-tissue CT windowing (window width 400 HU, window level 40 HU) and extracted 2D axial slices from the 3D volumes. We then converted segmentation masks to text-box pairs and excluded boxes below 0.01\% of the image area, yielding 4,489 pairs from 1,855 images for training, 586 from 236 images for validation, and 558 from 235 images for testing.

\subsubsection{Semantics-Centric Datasets}
Semantics-centric datasets target tasks that require image-level semantic understanding.
\textbf{SLAKE}~\cite{liu2021slake} consists of 642 medical images from CT, MRI, and X-ray modalities and 14,028 open- and closed-ended QA pairs (see Fig.~\ref{fig_dataset}(d)). We used only the English subset. We followed the official split, with 4,919 QA pairs from 450 images for training, 1,053 from 96 images for validation, and 1,061 from 96 images for testing.
\textbf{VQA-RAD}~\cite{lau2018dataset} consists of 315 radiology images from CT, MRI, and X-ray modalities and 3,515 clinician-generated QA pairs (see Fig.~\ref{fig_dataset}(e)). We used only Yes/No QA pairs. We adopted the split from~\cite{saab2024capabilities} to avoid the data leakage present in the official split, with 418 QA pairs from 103 images for training, 382 from 92 images for validation, and 392 from 99 images for testing.
\textbf{OmniMedVQA}~\cite{hu2024omnimedvqa} consists of 118,010 medical images and 127,995 QA pairs collected from 73 datasets across 12 imaging modalities (see Fig.~\ref{fig_dataset}(f)). We used the open access subset from CARES~\cite{xia2024cares}, which was selected to enrich rare modalities and anatomical regions underrepresented in other datasets. It spans mammography, endoscopy, fundus photography, hand X-ray, CT, MRI, and ultrasound, comprising 8,188 images and 8,797 multiple-choice QA pairs. We retained only the question and ground-truth answer for each pair, discarding the multiple-choice options, and excluded questions with a single possible answer to avoid evaluation bias. This yielded 4,953 QA pairs from 4,780 images for training, 663 from 637 images for validation, and 625 from 613 images for testing.

\begin{figure}[t]
    \centering
    \includegraphics[width=0.6\linewidth]{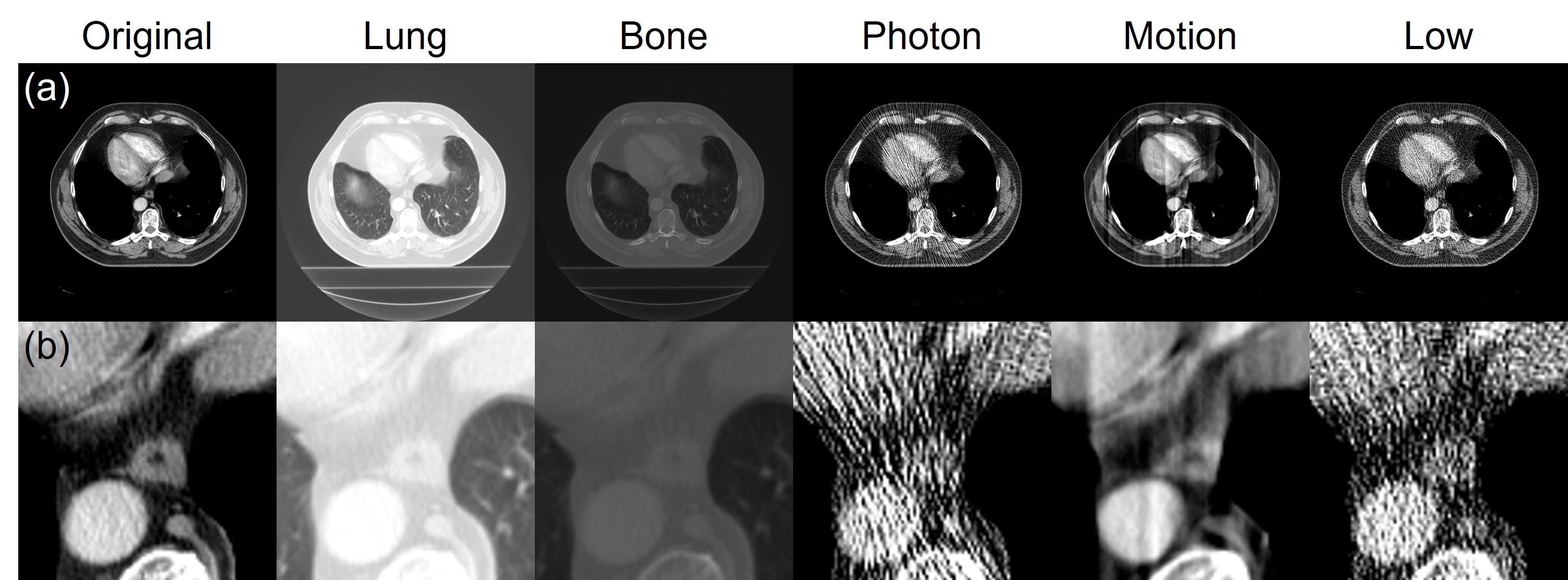}
    \caption{Examples of perturbed test images under five perturbation types: lung-window, bone-window, photon starvation, motion artifact, and low-dose degradation. (a) shows the whole test image and (b) the zoomed-in region.}
    \label{fig_perturbation}
\end{figure}

\subsubsection{Perturbation Dataset}
We constructed \textbf{SegTHOR-Cls} from SegTHOR for region-based organ classification, where the task is to predict the organ label for an image region specified by a bounding box. The dataset comprises training, validation, original test, and perturbed test splits, with the training, validation, and original test splits identical to those of SegTHOR. The perturbed test split includes five perturbation types: lung-window, bone-window, photon starvation, motion artifact, and low-dose degradation (see Fig.~\ref{fig_perturbation})~\cite{barrett2004artifacts}. Lung-window uses a conventional lung CT window setting (window width 1500 HU, window level -600 HU), whereas bone-window uses a conventional temporal bone CT window setting (window width 4000 HU, window level 700 HU). Photon starvation simulates severe streaking artifacts by reducing the effective photon budget for projection rays whose attenuation exceeds the 70th percentile of the attenuation distribution. Motion artifact simulates patient motion during acquisition by applying a 5 mm displacement to the sinogram, resulting in shading and streaking artifacts. Low-dose degradation is simulated by reducing the incident photon count to $1.0\times10^5$, increasing quantum noise in the reconstruction. All three reconstruction-based perturbations are simulated under a parallel-beam approximation of the clinical fan-beam geometry.

\subsection{Experimental Settings}
We trained our vision encoder on MedG to obtain LoFi, then compared it against two groups of foundation models: (1) general-purpose foundation models trained on natural images, including SigLIP2-400M~\cite{tschannen2025siglip} (SigLIP), DINOv3-Large~\cite{simeoni2025dinov3} (DINO), and V-JEPA 2.1-Large~\cite{mur2026v} (JEPA); and (2) medical vision foundation models trained on medical images, including MedSigLIP~\cite{sellergren2025medgemma}, RadioDINO~\cite{zedda2025radio}, and MedSAM~\cite{ma2024segment}.
For each downstream experiment, we selected the best checkpoint based on the validation metric and report the final results on the held-out test set. All experiments were repeated three times with different random seeds. For each vision encoder, we used the input resolution employed during its pretraining: DINO and JEPA were trained and evaluated at $512 \times 512$, while RadioDINO and MedSAM use lower and higher resolutions of $256 \times 256$ and $1024 \times 1024$, respectively. For the latter two, we adjusted the $\operatorname{AvgPool}$ configuration to align the spatial resolution of the patch embeddings for a comparable evaluation.
To assess the contribution of our location-aware objective, we trained the vision encoder with the grounding loss and the grounded captioning loss separately, and compared them against the full model trained with both. This ablation used SigLIP2-Base~\cite{tschannen2025siglip} with a batch size of 24, where one epoch corresponded to 500K sampled instances.

\subsection{Evaluation Metrics}
For phrase grounding, we report the F1-score at an intersection-over-union (IoU) threshold of 0.5 (F1@0.5), where precision (Pr@0.5) and recall (Re@0.5) are defined as $Pr = M/N_{\text{pred}}$ and $Re = M/N_{\text{gt}}$, respectively, and the F1-score is computed as $F1 = (2 \cdot Pr \cdot Re)/(Pr + Re)$. Here, $M$ denotes the number of matched prediction--ground-truth pairs with $\mathrm{IoU} \ge 0.5$, $N_{\text{pred}}$ is the total number of predicted boxes, and $N_{\text{gt}}$ is the total number of ground-truth boxes. This metric is more suitable for comparison than average precision, as many LVLMs do not provide reliable confidence scores~\cite{jiang2026detect}.
For both VQA and region-based organ classification, we report accuracy, defined as $\frac{1}{N_{\text{q}}}\sum_{i=1}^{N_{\text{q}}} s_i$, where $s_i$ is the matching score for question $i$ and $N_{\text{q}}$ is the total number of questions. The matching score $s_i$ equals 1 when the normalized predicted answer exactly matches the normalized ground-truth answer, and 0 otherwise, following~\cite{li2023llava}. Although the SLAKE and OmniMedVQA test sets contain 2 and 4 answers, respectively, that never appear during training, we regard this as an inherent property of open-ended question answering and adopt the same evaluation protocol.
For the perturbed test set, we compute the entropy $-\sum_{i} p_i \log_2 p_i$ over the probability distribution of the special label tokens, where $p_i$ is the normalized probability of the $i$-th special label token. To visualize how each perturbation redistributes the model's confidence, we present a heatmap of the mean change in normalized probability, $p_i^{\text{perturbed}} - p_i^{\text{original}}$, for each special label token, averaged over image pairs sharing the same ground-truth label.
\begin{figure}[t]
\centering
\begin{subfigure}{0.32\textwidth}
\centering
\includegraphics[width=\linewidth]{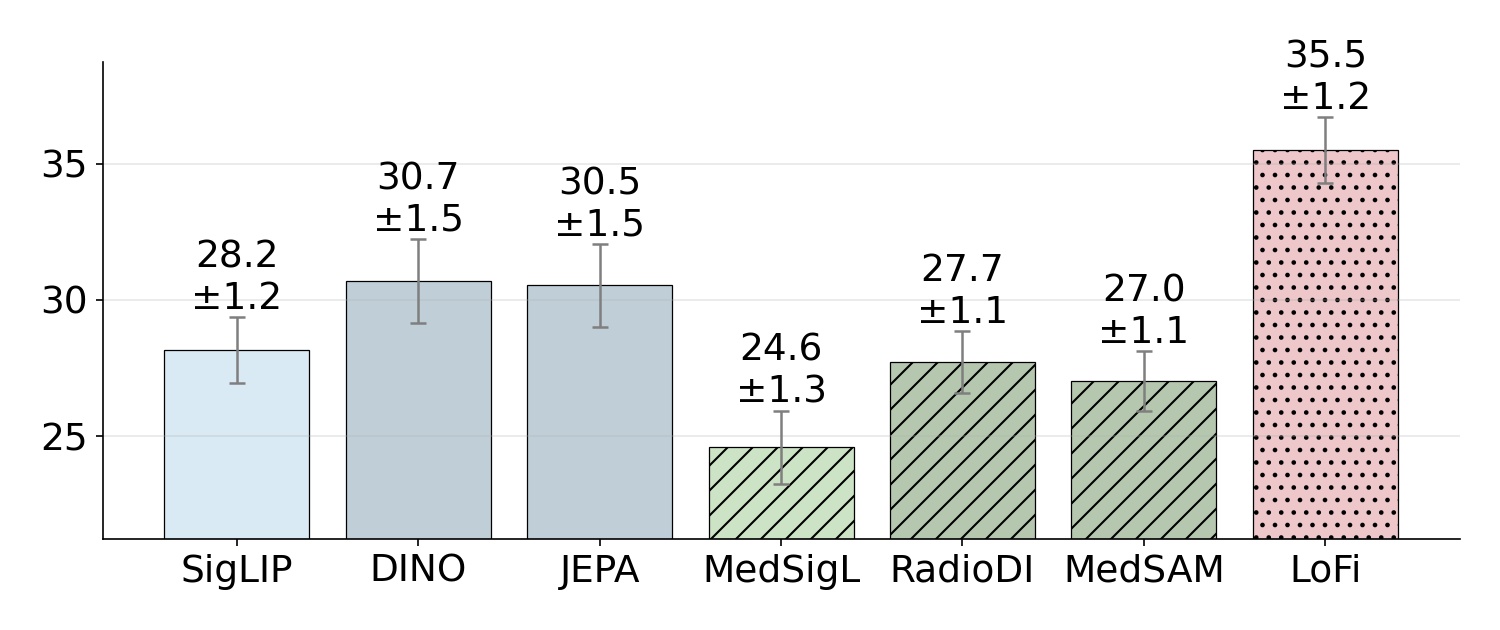}
\caption{PadChest-GR}
\end{subfigure}
\begin{subfigure}{0.32\textwidth}
\centering
\includegraphics[width=\linewidth]{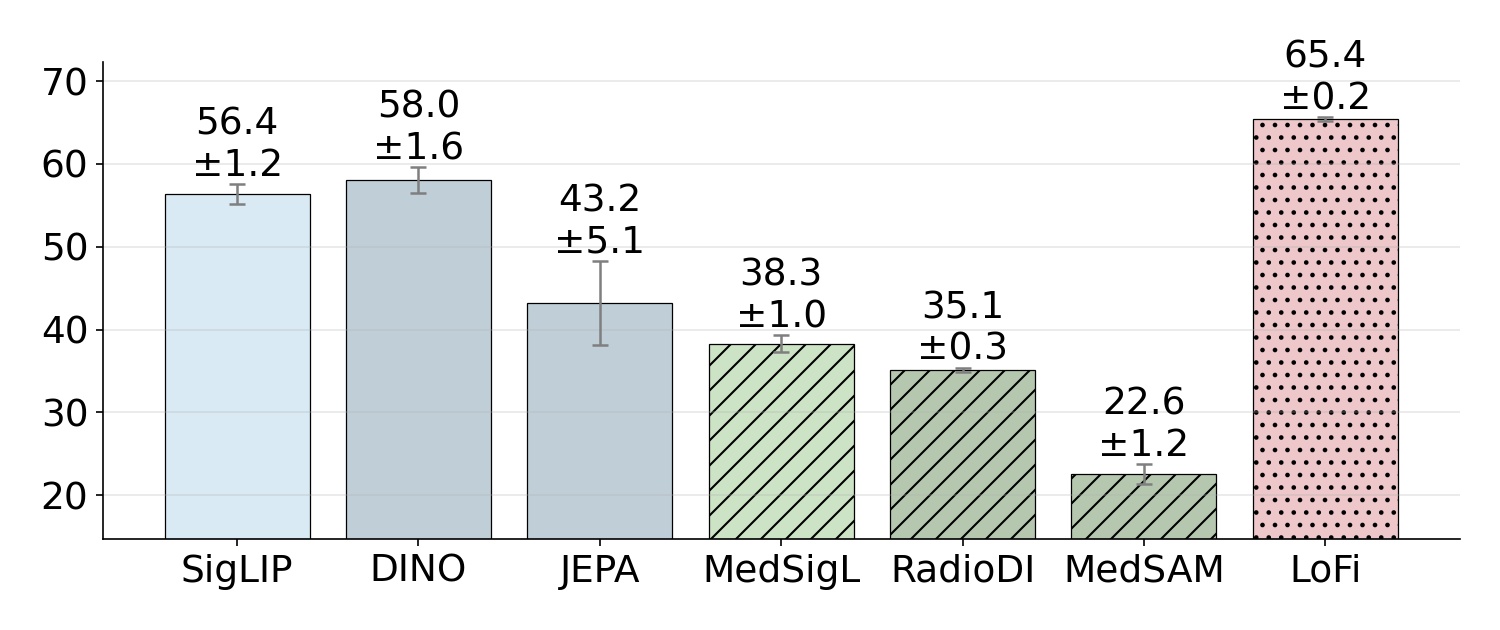}
\caption{TN5000}
\end{subfigure}
\begin{subfigure}{0.32\textwidth}
\centering
\includegraphics[width=\linewidth]{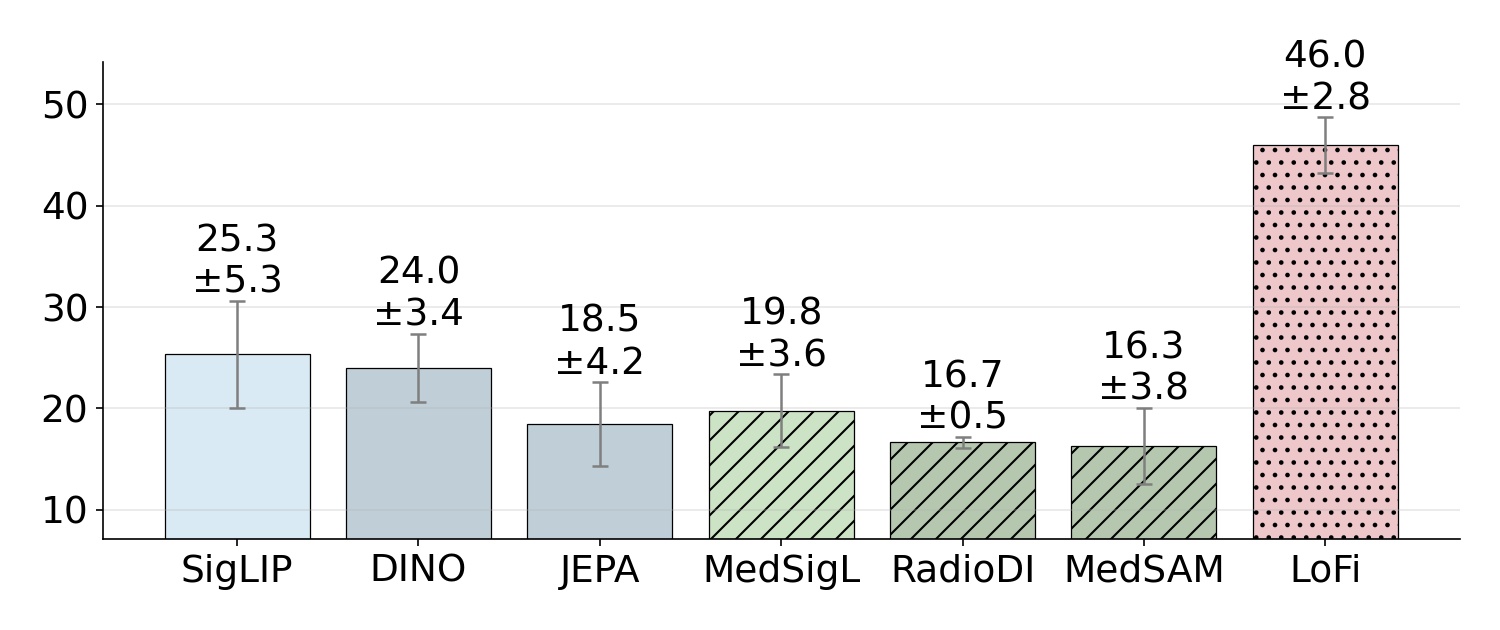}
\caption{SegTHOR}
\end{subfigure}
\begin{subfigure}{0.32\textwidth}
\centering
\includegraphics[width=\linewidth]{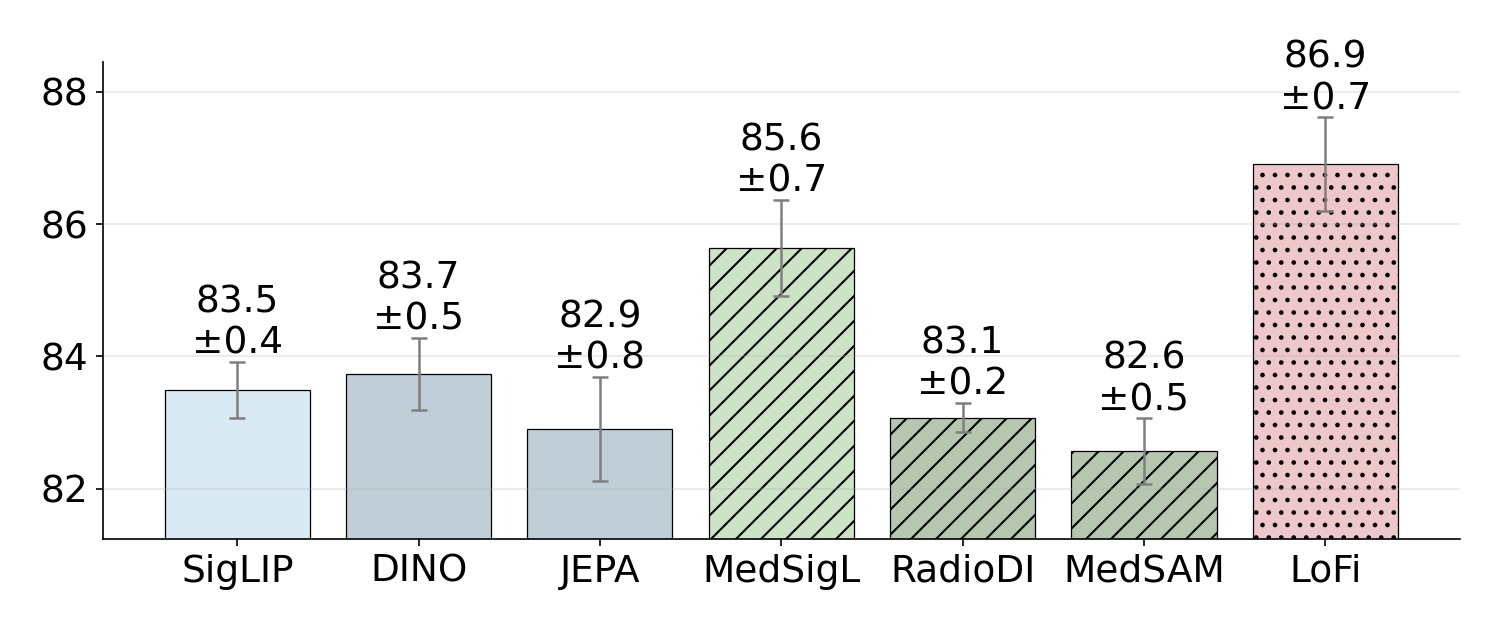}
\caption{SLAKE}
\end{subfigure}
\begin{subfigure}{0.32\textwidth}
\centering
\includegraphics[width=\linewidth]{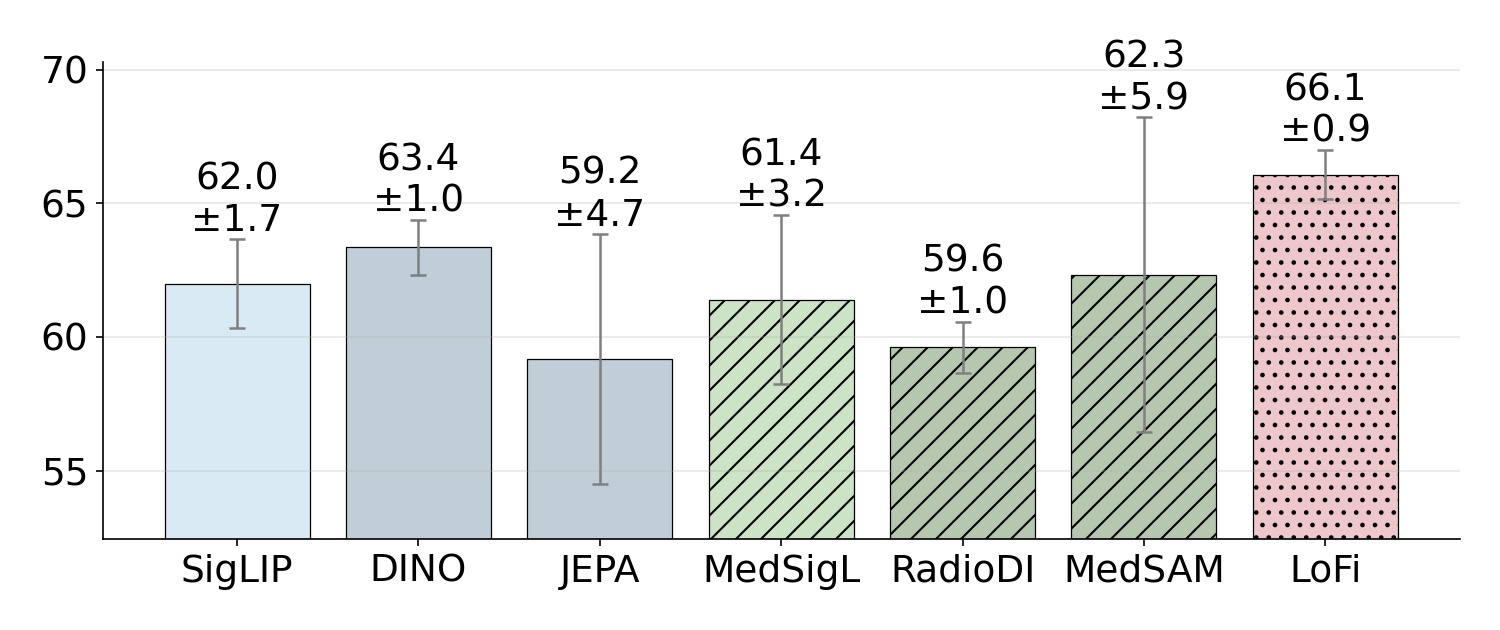}
\caption{VQA-RAD}
\end{subfigure}
\begin{subfigure}{0.32\textwidth}
\centering
\includegraphics[width=\linewidth]{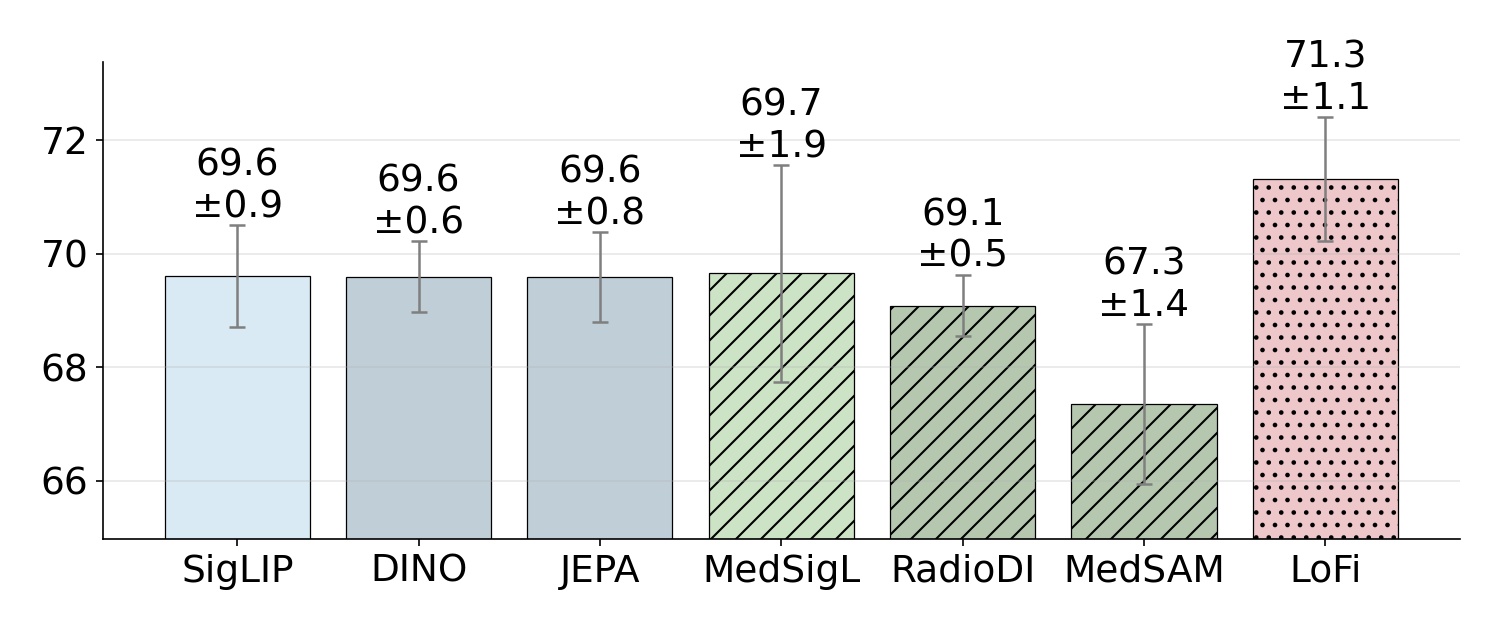}
\caption{OmniMedVQA}
\end{subfigure}
\caption{Results of downstream fine-tuning on perception-centric datasets (a)-(c) and semantics-centric datasets (d)-(f). We compared LoFi (red) with general-purpose foundation models (blue): SigLIP, DINO, and JEPA; and medical vision foundation models (green): MedSigLIP, RadioDINO, and MedSAM. Models trained without image-text alignment are rendered in darker tones. The y-axis shows F1@0.5 in (a)-(c) and accuracy in (d)-(f).}
\label{fig_main_results}
\end{figure}

\begin{figure}[t]
    \centering
    \includegraphics[width=0.75\linewidth]{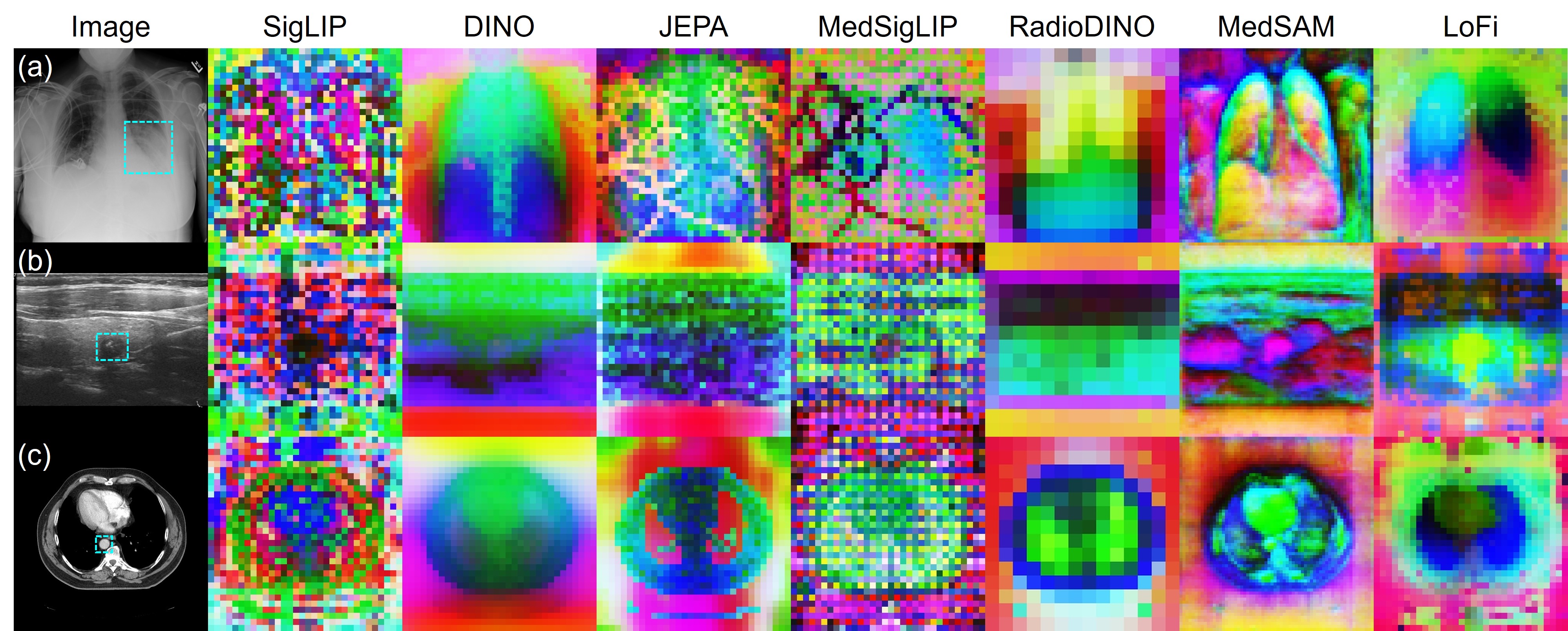}
    \caption{PCA maps of patch-level features from SigLIP, DINO, JEPA, MedSigLIP, RadioDINO, and MedSAM on (a) a chest X-ray from SLAKE, (b) an ultrasound image from TN5000, and (c) a CT slice from SegTHOR. The cyan box indicates the target ROI for phrase grounding.}
    \label{fig_pca_all}
\end{figure}

\section{Results}
\subsection{Perception-Centric Dataset Results}
Fig.~\ref{fig_main_results}(a)-(c) show the F1@0.5 scores on the PadChest-GR, TN5000, and SegTHOR datasets. LoFi outperforms all prior models across all datasets. We attribute this improvement primarily to the large-scale grounding dataset and the location-aware learning strategy, which together encourage clinically meaningful and spatially consistent patch-level features. Fig.~\ref{fig_pca_all} provides qualitative support: LoFi's patch-level features are highly consistent within anatomical structures and effectively separate small targets (e.g., the aorta) from surrounding structures (e.g., the heart).
We next examine the behavior of prior models on each dataset. On PadChest-GR, DINO and JEPA outperform medical-domain models such as MedSigLIP. This is consistent with the visualizations in Fig.~\ref{fig_pca_all}(a), where they produce features with high spatial consistency.
On TN5000, DINO ranks first among prior models, followed by SigLIP. SigLIP's advantage over JEPA is evident in Fig.~\ref{fig_pca_all}(b), where the nodule ROI is more clearly distinguishable. We attribute this to SigLIP's auxiliary objectives, which promote spatial consistency and yield more discriminative nodule features.
On SegTHOR, SigLIP ranks first, followed by DINO; as shown in Fig.~\ref{fig_pca_all}(c), DINO's features struggle to distinguish small organs (e.g., the aorta) from adjacent structures, leading to lower performance.
Interestingly, medical vision foundation models fail to outperform their general-purpose counterparts despite domain-specific pretraining. As shown in Fig.~\ref{fig_pca_all}, MedSigLIP is often distracted by spurious objects (e.g., tubes), while RadioDINO and MedSAM fail to separate target organs from neighboring non-target structures, resulting in inaccurate box predictions.

\subsection{Semantics-Centric Dataset Results}
Fig.~\ref{fig_main_results}(d)-(f) show the accuracy on the SLAKE, VQA-RAD, and OmniMedVQA datasets. 
LoFi achieves the best performance across all datasets. Unlike models that enforce spatial consistency through self-supervised objectives (e.g., DINO), LoFi lets this consistency emerge from clinically grounded objectives, yielding features that are not only spatially consistent but also clinically meaningful. These gains come from the learning objective rather than data scale: despite training on comparable or less data, LoFi (1.37M images) outperforms both MedSigLIP (33M image-text pairs) and RadioDINO (1.35M images).
The best-performing prior model, however, varies by dataset. On SLAKE, MedSigLIP surpasses all other prior models, indicating that image-text alignment training on well-curated medical datasets benefits semantics-centric tasks. This contrasts with our findings on perception-centric tasks, where self-supervised models performed best. 
On VQA-RAD, DINO ranks first; because VQA-RAD contains far fewer training samples (418) than SLAKE (4,919), DINO's advantage may stem from its less noisy, more homogeneous patch-level features, which reduce reliance on spurious cues during training.
On OmniMedVQA, MedSigLIP again ranks first, but the margins are modest. This likely reflects its limited ability to extract discriminative features from rare imaging modalities, which yields similar performance across models.

\begin{figure}[t]
\centering
\begin{subfigure}{1.0\linewidth}
\centering
\includegraphics[width=\linewidth]{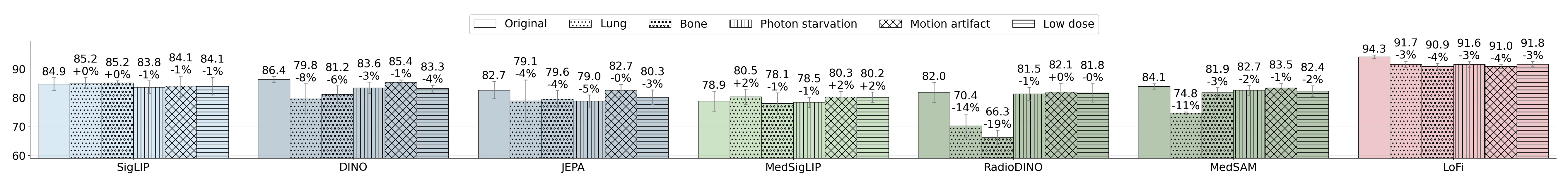}
\caption{Accuracy and relative accuracy change (\%) compared to the original}
\end{subfigure}
\begin{subfigure}[t]{0.22\linewidth}
\raggedright
\includegraphics[width=\linewidth]{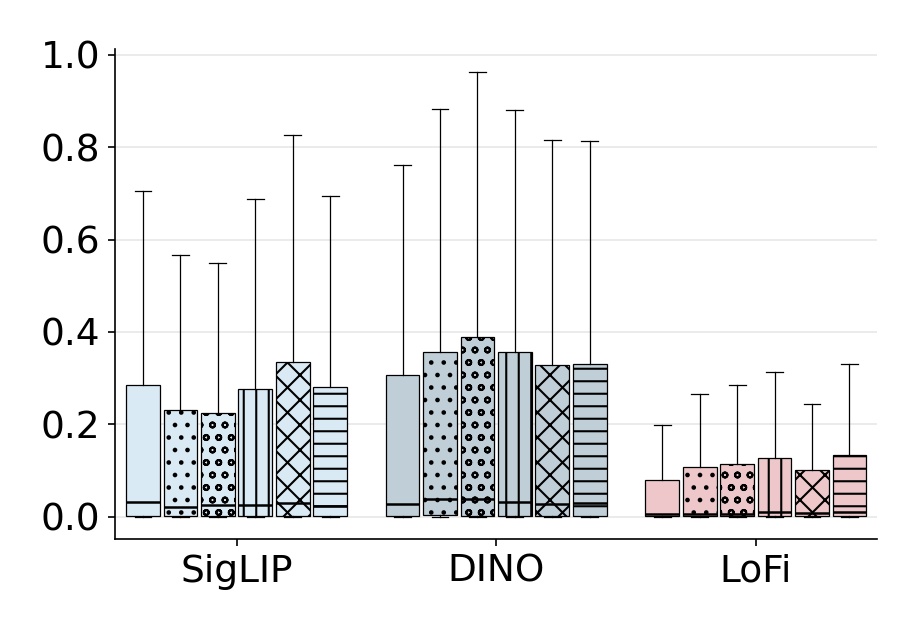}
\caption{Entropy}
\end{subfigure}
\begin{subfigure}[t]{0.47\linewidth}
\centering
\includegraphics[width=\linewidth]{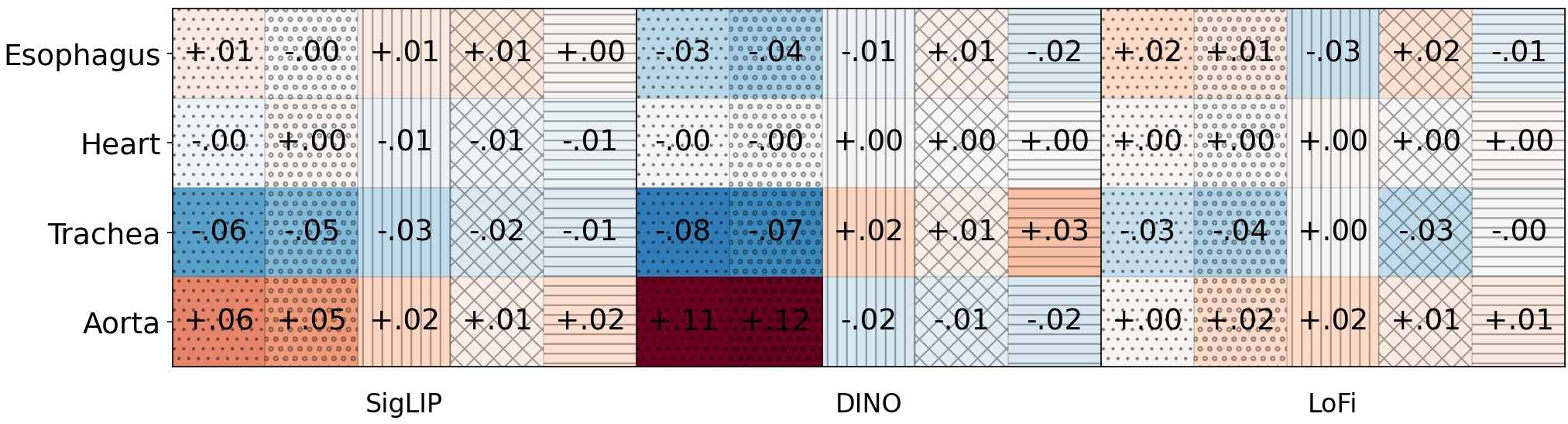}
\caption{Mean change in probability from original}
\end{subfigure}
\begin{subfigure}[t]{0.29\linewidth}
\raggedleft
\includegraphics[width=\linewidth]{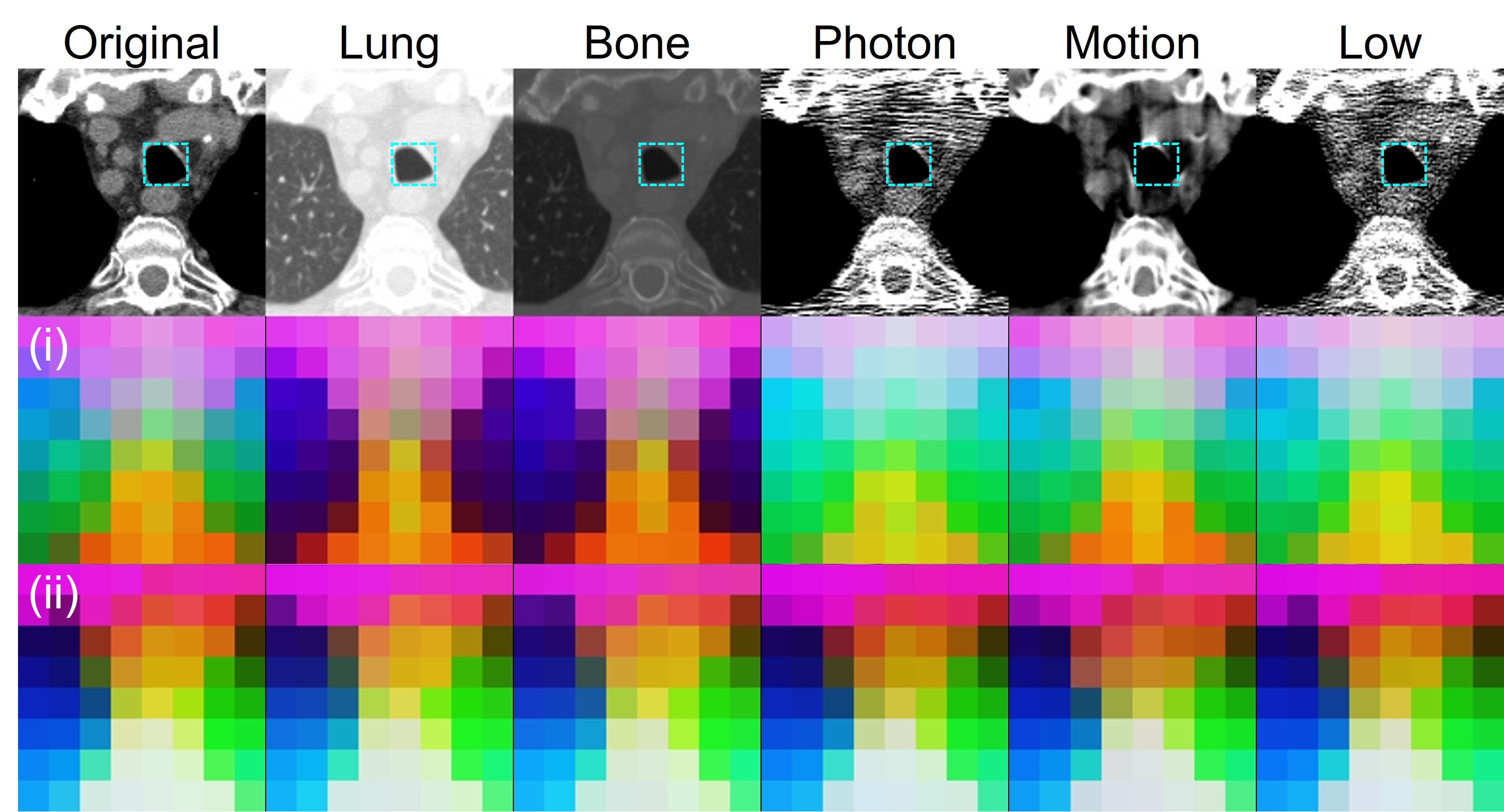}
\caption{PCA maps}
\end{subfigure}
\caption{Results of region-based organ classification on the SegTHOR-Cls dataset. We report (a) accuracy on the original test set and five perturbed test sets; (b) entropy values for four special label tokens from the top-3 most accurate models; (c) a heatmap of the mean change in probability from the original to the perturbed test set for the same top-3 models; and (d) PCA maps of patch-level features from (i) DINO and (ii) LoFi. In (d), aggregated patch-level features from cropped regions of the original and perturbed test sets are projected onto three dimensions via PCA. The cyan box indicates the trachea.}
\label{fig_perturbation_results}
\end{figure}

\begin{figure}[t]
\centering
\begin{subfigure}{0.24\linewidth}
\centering
\includegraphics[width=\linewidth]{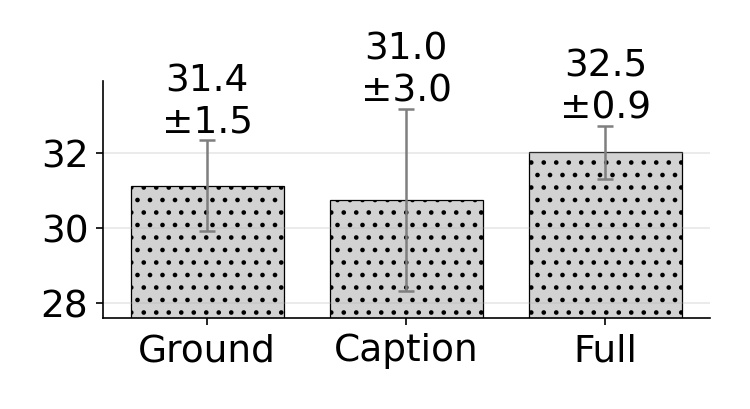}
\caption{PadChest-GR}
\end{subfigure}
\begin{subfigure}{0.24\linewidth}
\centering
\includegraphics[width=\linewidth]{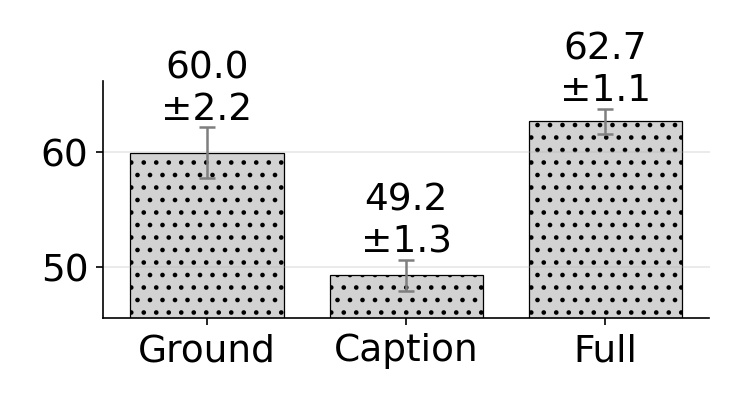}
\caption{TN5000}
\end{subfigure}
\begin{subfigure}{0.24\linewidth}
\centering
\includegraphics[width=\linewidth]{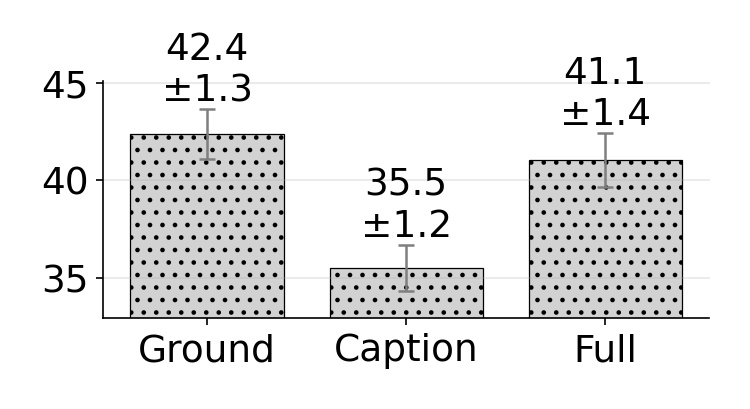}
\caption{SegTHOR}
\end{subfigure}

\begin{subfigure}{0.24\linewidth}
\centering
\includegraphics[width=\linewidth]{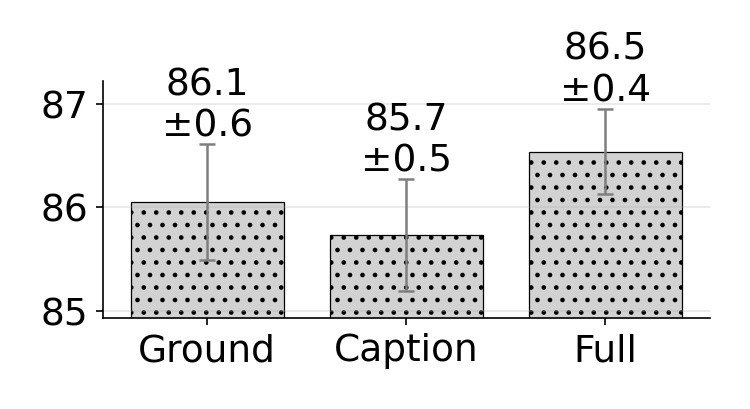}
\caption{SLAKE}
\end{subfigure}
\begin{subfigure}{0.24\linewidth}
\centering
\includegraphics[width=\linewidth]{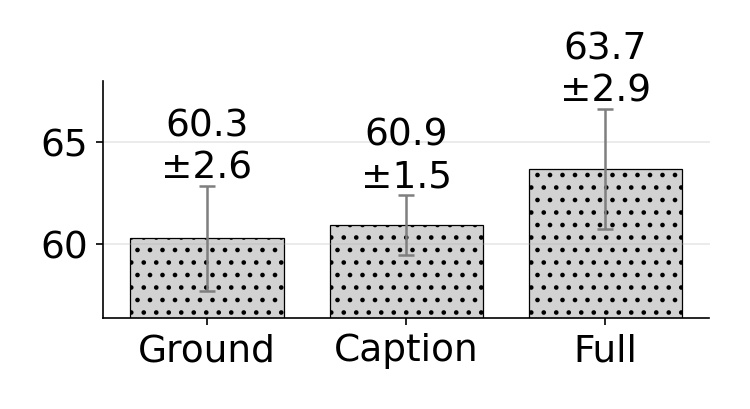}
\caption{VQA-RAD}
\end{subfigure}
\begin{subfigure}{0.24\linewidth}
\centering
\includegraphics[width=\linewidth]{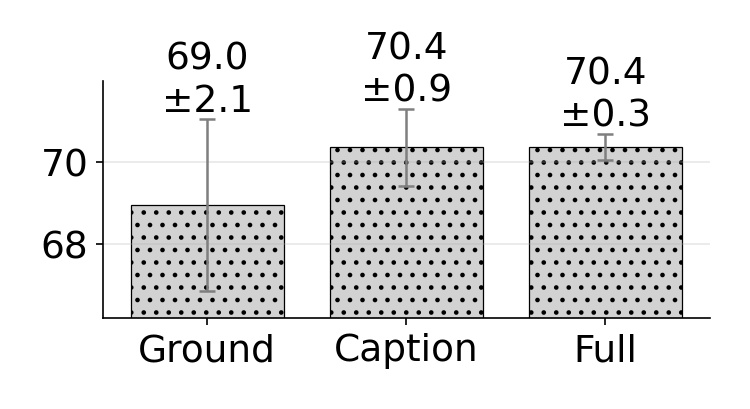}
\caption{OmniMedVQA}
\end{subfigure}
\caption{Ablation on each training objective in LoFi with SigLIP2-Base, comparing grounding loss only (Ground), grounded captioning loss only (Caption), and both jointly (Full).}
\label{fig_ablation}
\end{figure}

\subsection{Perturbation Dataset Results}
Self-supervised encoders such as DINO show strong spatial consistency on the perception-centric tasks, but this robustness comes from enforcing invariance to a fixed set of training augmentations, leaving them vulnerable to the out-of-distribution perturbations in Fig.~\ref{fig_perturbation}. LoFi, by contrast, does not rely on such augmentations; it is instead trained to align clinical text with bounding boxes. We examine how this distinction manifests under perturbation.
Fig.~\ref{fig_perturbation_results}(a) reports accuracy on SegTHOR-Cls. On the original test set, LoFi ranks first and DINO second, consistent with the perception-centric results. LoFi degrades only slightly under perturbation and still outperforms all competing models on both the original and perturbed sets, achieving the highest absolute accuracy while degrading less than DINO.
Fig.~\ref{fig_perturbation_results}(b) shows the entropy of the special label tokens. LoFi attains the lowest entropy among the top-3 models together with the highest accuracy, indicating that its patch-level features are more discriminative than those of competing methods, effectively separating an organ from its neighbors. DINO, ranked second, shows markedly higher entropy under the lung- and bone-window perturbations. As Fig.~\ref{fig_perturbation} illustrates, these perturbations brighten air regions that appear black in the original images, lowering the mean trachea probability while raising the mean aorta probability (Fig.~\ref{fig_perturbation_results}(c)); this aggravates confusion between nearby organs and reduces accuracy. SigLIP exhibits clear probability shifts under the same perturbations, and although its relative degradation is smaller, its absolute accuracy remains lower. Fig.~\ref{fig_perturbation_results}(d) traces these probability changes to the features themselves, where DINO shows substantially greater feature drift than LoFi under the lung- and bone-window perturbations.
Overall, the invariances learned by DINO do not fully cover these clinically motivated perturbations, whereas our location-aware objective yields spatial consistency that degrades comparatively less under them.

\subsection{Ablation Study}
Fig.~\ref{fig_ablation} presents the ablation results for the two training objectives in LoFi. Using the grounded captioning loss alone yields clearly lower performance on the perception-centric tasks (a)-(c), whereas using the grounding loss alone tends to underperform on the semantics-centric tasks (i.e., VQA-RAD and OmniMedVQA). Combining both objectives achieves the best or comparable results across both task types, with SegTHOR being the only exception. Overall, these results demonstrate that the two losses are complementary.

\subsection{Comparison with SOTA Methods}
We evaluated our fine-tuned model against SOTA LVLMs specialized for each benchmark's task and domain. On PadChest-GR, we compared Pr@0.5, Re@0.5, and F1@0.5 with chest X-ray LVLMs that support phrase grounding: CheXagent~\cite{chen2024chexagent}, RadVLM~\cite{deperrois2025radvlm}, and MAIRA-2~\cite{bannur2024maira}. On SLAKE, we compared accuracy with medical VQA LVLMs: Fleming-VL-8B~\cite{shu2025fleming} (Fleming), an open-source medical LVLM, and MedMO-8B-Next~\cite{deria2026medmo} (MedMO), which leverages bounding boxes during training. All baselines were trained on the training split of each downstream dataset. Following MedMO and Fleming, we evaluated VQA accuracy with an LLM-as-judge protocol using \texttt{gpt-5-mini-2025-08-07} for direct comparability.

Table~\ref{tab_comp_padchest} reports grounding performance on PadChest-GR. LoFi attains a mean F1@0.5 of 35.5, outperforming all competing methods. Among the compared LVLMs, RadVLM underperforms MAIRA-2 despite its more recent LLM backbone and its training on a larger chest X-ray dataset. We attribute this gap to their vision encoders: MAIRA-2 uses a vision encoder fine-tuned with a self-distillation objective on chest X-rays, whereas RadVLM fine-tunes a general-purpose LVLM. Although MAIRA-2 attains higher Pr@0.5, its conservative bounding box predictions yield lower Re@0.5, resulting in an F1@0.5 of 32.9.
Table~\ref{tab_comp_slake} reports VQA accuracy (LLM-as-judge) on SLAKE. LoFi achieves the best performance in both settings, with 88.5\% on closed-ended questions and 84.9\% overall. Under this shared SLAKE fine-tuning protocol, the comparison provides a quantitative measure of LoFi's competitiveness with SOTA LVLMs.

\begin{table}[t]
\centering
\caption{Comparison of LoFi against CheXagent, MAIRA-2, and RadVLM on the PadChest-GR dataset.}
\label{tab_comp_padchest}
\begin{tabular}{llll}
\hline
Model & Pr@0.5 & Re@0.5 & F1@0.5 \\
\hline
CheXagent & 20.8 & 7.7 & 11.3 \\
MAIRA-2 & 40.0 & 27.9 & 32.9 \\
RadVLM$^\ast$ & 27.7 & 31.0 & 29.2 \\
LoFi & 35.8 ($\pm1.3$) & 35.2 ($\pm1.1$) & \textbf{35.5 ($\pm1.2$)} \\
\hline
\multicolumn{4}{l}{\scriptsize * indicates test data seen during training.} \\
\end{tabular}
\end{table}

\begin{table}[t]
\centering
\caption{Comparison of LoFi against Fleming and MedMO on the SLAKE dataset.}
\label{tab_comp_slake}
\begin{tabular}{lll}
\hline
Model & Closed & All \\
\hline
Fleming & 86.9 & 80.0 \\
MedMO & 83.0 & 81.6 \\
LoFi (LLM-as-judge) & \textbf{88.5 ($\pm0.8$)} & \textbf{84.9 ($\pm0.7$)} \\
\hline
\end{tabular}
\end{table}

\begin{table}[t]
\centering
\caption{External validation of LoFi (w/o FT) against MedROV on the SegTHOR dataset.}
\label{tab_comp_segthor}
\begin{tabular}{lccc}
\hline
Model & Pr@0.5 & Re@0.5 & F1@0.5 \\
\hline
MedROV & 20.8 & 9.2 & 12.7 \\
LoFi (w/o FT) & 41.2 & 44.2 & \textbf{42.6} \\
\hline
\end{tabular}
\end{table}

\subsection{External Validation on Grounding}
We compared the grounding capability of LoFi against MedROV~\cite{sheikh2026medrov}, a SOTA medical open-vocabulary detection model trained on over 600K samples. Since neither model was trained on SegTHOR, this comparison assesses grounding transfer to an unseen dataset. We denote by LoFi (w/o FT) the direct evaluation on the SegTHOR test set without fine-tuning, and report Pr@0.5, Re@0.5, and F1@0.5. For MedROV, we computed phrase-grounding performance by aggregating all predicted boxes with confidence above 0.001 for each ground-truth label and comparing them against the corresponding ground-truth box.

Table~\ref{tab_comp_segthor} shows the grounding performance of LoFi (w/o FT) and MedROV on SegTHOR. LoFi (w/o FT) outperforms MedROV by a large margin in F1@0.5, demonstrating superior grounding capability relative to this SOTA model.
\section{Conclusion}
In this paper, we presented LoFi, a medical vision foundation model built on location-aware fine-grained representation learning. We observed that no existing medical vision foundation model produces patch-level features that are both clinically meaningful and spatially consistent, and that instruction-tuned LVLMs largely inherit the patch-level properties of their vision encoders. Motivated by this, we incorporated location information into vision encoder training, upstream of the language model. We constructed MedG, a large-scale medical grounding dataset of 4.48M image-text-box triplets aggregated from 84 datasets across 7 modalities, and trained the vision encoder with a lightweight LLM using grounding and grounded captioning objectives, letting spatial consistency emerge without any direct patch-level regularization. Extensive experiments on perception-centric phrase grounding, semantics-centric VQA, and a perturbation benchmark show that LoFi consistently outperforms both general-purpose and medical vision foundation models as well as SOTA LVLMs. We hope this work underscores the importance of strengthening the visual foundation upstream of the language model and encourages further research on fine-grained representation learning for reliable medical LVLMs.

\section*{Acknowledgments}
The research is supported, in part, by the NSERC Discovery Grant RGPIN-2022-05316, NSERC Alliance Grant ALLRP 602633-24, Tri-Agency Canada; Canada CIFAR AI Chair Awards and Canada Research Chair Fellowship; Genome British Columbia (ISI003), the Terry Fox Research Institute, Health Research BC; Google Gemini Research Awards; IITP grant, the Ministry of Science and ICT (RS-2024-00445087, RS-2025-25464461), funded by the Korea government (MSIT); the National Research Foundation of Korea (NRF) grant funded by the Korea government (MSIT) (RS-2025-00515536).

\bibliographystyle{plainnat}
\bibliography{refs} 

\end{document}